\documentclass{article}

\usepackage{arxiv}

\usepackage[utf8]{inputenc} 
\usepackage[T1]{fontenc}    
\usepackage{hyperref}       
\usepackage{url}            
\usepackage{booktabs}       
\usepackage{amsfonts}       
\usepackage{nicefrac}       
\usepackage{microtype}      
\usepackage{lipsum}		
\usepackage{graphicx}
\usepackage{doi}
\usepackage{comment}
\usepackage{amsfonts}       
\usepackage{amsmath}
\usepackage{mathrsfs}
\usepackage[noend]{algpseudocode}
\algnewcommand{\LineComment}[1]{\State \(\triangleright\) #1}
\usepackage{algorithmicx}
\usepackage{algorithm}
\usepackage{subcaption }

\title{FAST BLOCK LINEAR SYSTEM SOLVER USING Q-LEARNING SCHEDULING FOR UNIFIED DYNAMIC POWER SYSTEM SIMULATIONS}

\renewcommand{\shorttitle}{\textit{arXiv} Template}

\author{
 Yingshi Chen   \\
Giga Design Automation Co., Ltd\\
Shenzhen 518055, China\\
\texttt{yschen@giga.com.cn} \\

   \And
 Xinli Song \thanks{Co-first authors with equal contribution} , HanYang Dai, Tao Liu, Wuzhi Zhong, Guoyang Wu \\
  Power system Dept.\\
  Electric Power Research Institute\\
  Beijing, 100192, China\\ 
  \texttt{songxl@epri.sgcc.com.cn} \\
  
}

\begin{document}
\maketitle

\begin{abstract}
 We present a fast block direct solver for the unified dynamic simulations of power systems. This solver uses a novel Q-learning based method for task scheduling. Unified dynamic simulations of power systems represent a method in which the electric-mechanical transient, medium-term and long-term dynamic phenomena are organically united. Due to the high rank and large numbers in solving, fast solution of these equations is the key to speeding up the simulation. The sparse systems of simulation contain  complex nested block structure, which could be used by the solver to speed up. For the scheduling of blocks and frontals in the solver, we use a learning based task-tree scheduling technique in the framework of Markov Decision Process. That is, we could learn optimal scheduling strategies by offline training on many sample matrices. Then for any systems, the solver would get optimal task partition and scheduling on the learned model. Our learning-based algorithm could help improve the performance of sparse solver, which has been verified in some numerical experiments. The simulation on some large power systems shows that our solver is 2-6 times faster than KLU, which is the state-of-the-art sparse solver for circuit simulation problems.
\end{abstract}


\section{Introduction}
Unified dynamic simulations of power systems represent a method in which the electric-mechanical transient, medium-term and long-term dynamic phenomena are organically united. Facilitated by the use of implicit integral algorithms, they can simulate fast as well as slow phenomena in the long-term dynamic courses. During the simulation process, it is a one of key steps to solve sparse systems composed of two kinds of linear equations: difference equations from dynamic equipments and algebraic equations from power network. Due to the high rank and large numbers in solving, fast solution of these equations play an important role in unified dynamic simulations.

To speed up the simulation, a parallel algorithm of large linear system solver based on block matrix is proposed, in which a learning-based scheduler is used for parallel programming. This scheduler algorithm is  actually derived from the Markov Decision Process (MDP in short) \cite{bellman1957markovian,gosavi2015solving,chen2021learning}. MDP based learning methods have recently been in great glory in academia and industry. For example, Google's AlphaGO\cite{silver2016mastering} defeated the go master Lee Sedol. The engine of AlphaGO is based on deep reinforcement learning, whose theoretical basis is MDP. It's time to use these powerful modern weapons to improve the solution of large power systems.

\section{Block Structure of Power System Equations} \label{sec:solver}
As shown in Fig.1, the overall structure of sparse matrix in linear equations can be divided into 4 big sub-matrices. The sub-matrix A and D is formed by the equations from dynamic equipments and power networks respectively. As the coupling parts between A and D, the sub-matrix B and C represents the influences from power network and dynamic equipments respectively, i.e., bus voltages and input currents.
\begin{figure}[!ht]
\begin{center}
\includegraphics[width=0.6\textwidth]{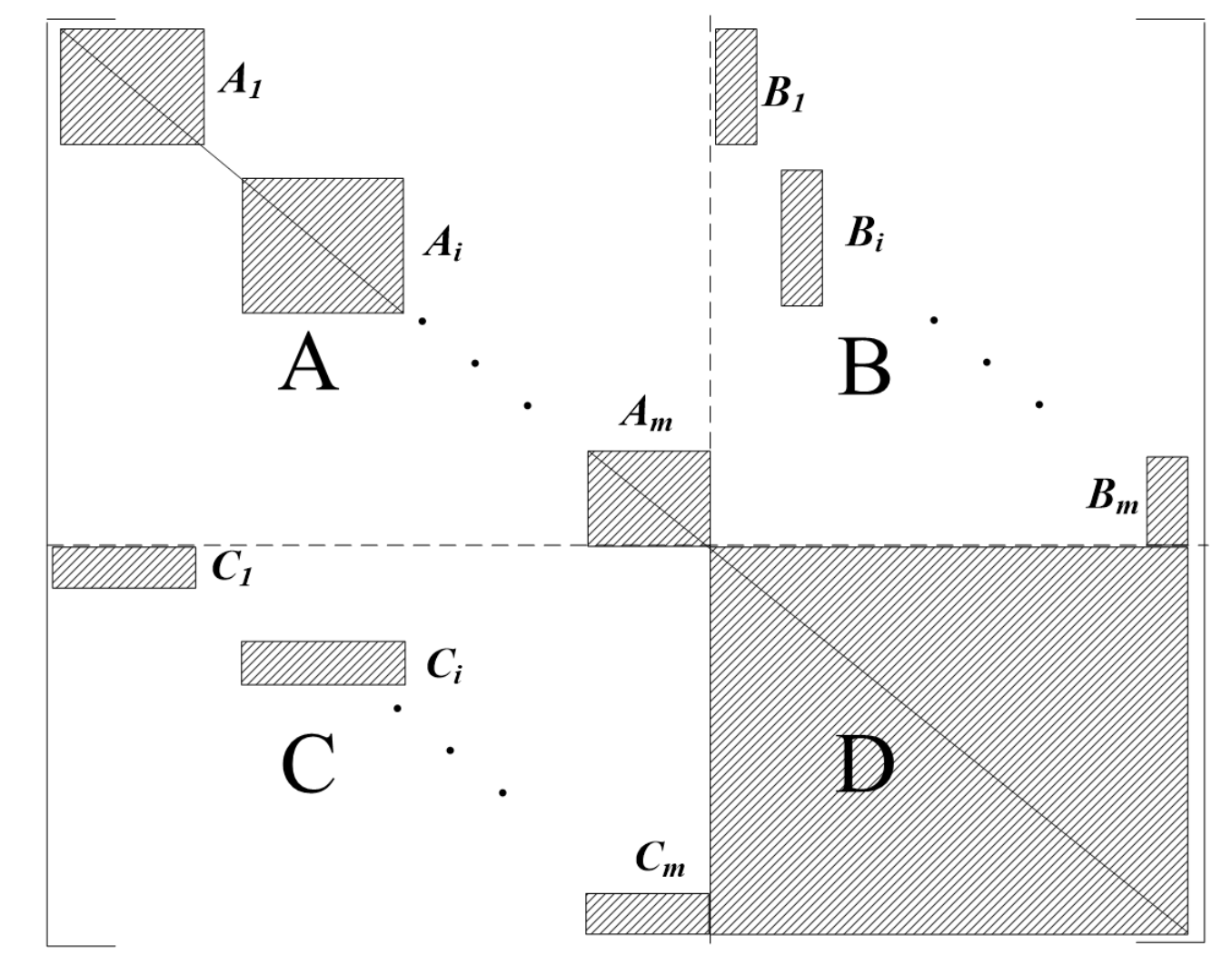}
\end{center}
\caption[position=bottom]{  Overall structure of sparse matrix from dynamic simulations of power systems }
\end{figure}

The rank of sub-matrix A is 3-5 times larger that of sub-matrix D. Sub-matrix B and C have a few elements. Therefore, most computation time is spent on A and D. Because both A and D have good structures for parallelism, we can get good speed-up. Like other general direct sparse solvers, this solver includes 3 phases, which are symbolic analysis, numeric factorization (LU decomposition) and solve (forward and backward substitution).
 
 For sub-matrix A, a new block-reuse parallel method is proposed, which is based on the facts that A is composed of many small independent diagonal blocks $A_i$, and many blocks possess the same symbolic structure, such as those buses on which the same type of loads or generators are located. 

\section{Q-Learning based Adaptive Scheduler} \label{sec:Scheduler}
In this section, we first introduce the background of Q-Learning, which is a powerful technique from Markov Decision Process and Reinforcement Learning. Then give the detail of Q-Learning based scheduler for the sparse solver.
\subsection{Q-Learning technique}

\subsubsection{Markov Decision Process (MDP)}    \label{sec:MDP}
The Markov decision process (MDP in short) \cite{bellman1957markovian,gosavi2015solving} is a smart and powerful model. As its name suggests, this model could find the solution (decision process) on the Markov property. The standard MDP includes four elements $\left ( E_n,O,P_o,R_o \right )$\cite{bellman1957markovian}:
\begin{itemize}
\item $E_n$ is the state space of MDP.
\item $O$ is all actions. We also use $O_s$ to denote the set of actions available from state $s$,
\item ${P_{o}(s,s')=\Pr(s_{t+1}=s'\mid s_{t}=s,o_{t}=o)}$ is the probability that action $o$ in state $s$ at time $t$,
\item $R_{o}(s,s')$ is the immediate reward (or expected immediate reward) received after transitioning from state $s$ to state $s'$, due to action $o$.
\end{itemize}

A transition matrix $M$ is a square matrix used to describe the transitions of state space. 

\begin{equation}
M=
\begin{bmatrix} 
	M_{1,1} & \cdots & M_{1,j} & \cdots & M_{1,n} \\
	& & \vdots   \\
	M_{i,1} & \cdots & M_{i,j} & \cdots & M_{i,n} \\
	& & \vdots   \\
	M_{n,1} & \cdots & M_{n,j} & \cdots & M_{n,n} \\
\end{bmatrix}
\end{equation}

Each element $ M_{i,j} $ is a non-negative real number representing a probability $ Pr(j|i) $ of moving from state $i$ to state $j$.
Each row summing of $M$ is 1: $ \sum_{j} M_{i,j}=1 $.

MDP is the theoretical foundation of Reinforcement Learning\cite{kaelbling1996reinforcement}. A widely-used Reinforcement Learning technique is Q-Learning.

\subsubsection{Q-Learning}             \label{sec:Q}
The MDP model presents a general paradigm and an abstract mathematical framework. For practical problems, there are many powerful techniques, such as Q-Learning\cite{watkins1989learning,watkins1992q}. "Q" refers to the expected rewards for an action taken in a given state\cite{watkins1989learning}. At each step, Q-Learning would take some action on the estimate of Q-values. If we know "Q" value of any action in a state, then it's simple to find the optimal solution of MDP. 

The following is a typical framework of Q-Learning\cite{sutton2018reinforcement}.
\begin{algorithm}[H]
\caption{A general framework of Q-Learning} \label{alg:Q}
\begin{flushleft}
\textbf{Init Parameters:} 
Learning rate $\alpha \in \left ( 0,1 \right ]$, discount rate $\gamma $ and reward function $R$
\end{flushleft}

\begin{algorithmic} [1]
\State Initialize $Q\left ( s,a \right )$ for all state $s$ and action $a$. For the terminal state, $Q(terminal,\cdot)=0$
\For{ each episode}
    \State Pick a initial state $s$
    \While {true}
        \State Choose action $a$ for the current state $s$, using policy derived from $Q$ 
        \State Take action $a$ from some policy $\pi$ then enter the next state $s'$
        \State Get reward $r = R\left ( s,a \right )$
        \State Update Q by some method, for example:
        \State \quad $Q\left ( s,a \right ) \leftarrow Q\left ( s,a \right )+\alpha \left [ r+\gamma Q\left ( s',a' \right )-Q\left ( s,a \right )  \right ]$
        \State Set $s=s'$
        \If {$s$ is terminal}
            \State break
        \EndIf
    \EndWhile
\EndFor
\end{algorithmic}
\end{algorithm}
This framework shows a key advantage of Q-Learning - avoiding the usage of transition matrix. In the case of sparse solver, the probability in the transition matrix is hard to estimate. Or the transition matrix needs huge memory
The number of states of a million-order matrix is astronomical!). Instead, we could always get the reward or Q-value. So in practice, Q-Learning is more suitable for studying various combinatorial problems in GE. Watkins\cite{watkins1992q} prove that it would converge to the optimum Q-values with probability 1. So it's a reliable technique for the combinatorial optimization problems in sparse solver. 

Solution time is one of the most important metrics of sparse solver. We could avoid the training time in Q-Learning by offline technique. That is, first collect many matrices and train the model on these sample matrices. Then apply the learned Q-value to solve a matrix. This offline Q-Learning (or batch Q-Learning) has recently regained popularity as a viable path towards effective real-world application. For example, Conservative Q-learning\cite{kumar2020conservative,levine2020offline}.

\subsection{Q-Learning based task tree scheduling technique}
\begin{figure}[h]
\begin{tabular}{ll}
\quad \includegraphics[scale=0.3]{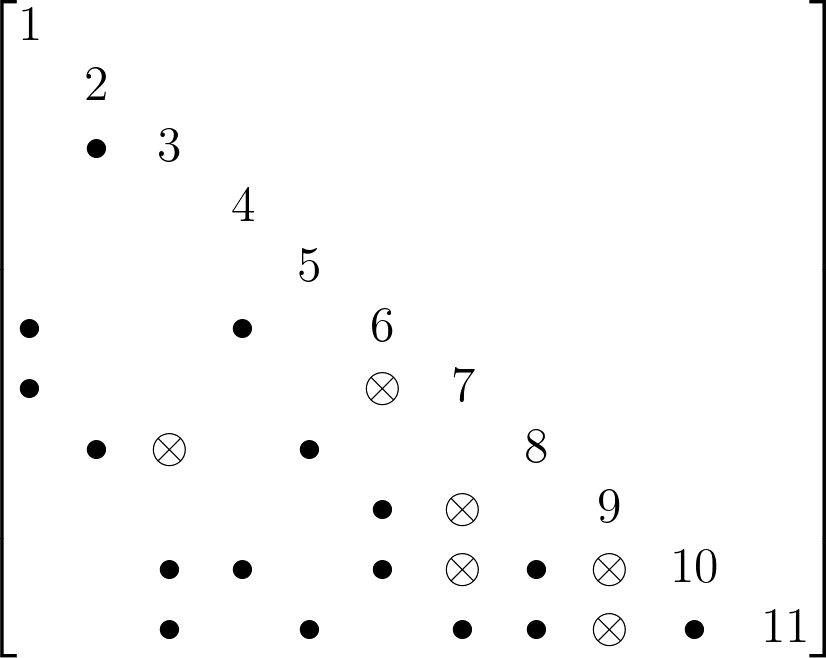}
&  \quad \quad \quad
\includegraphics[scale=0.3]{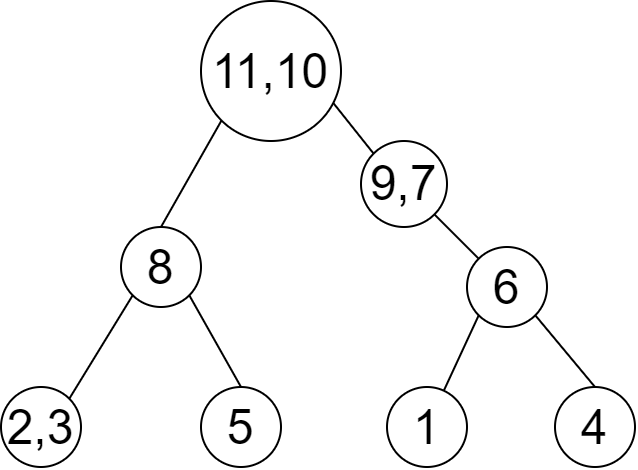}
\end{tabular}
\caption{(a) Pattern of a $11\times11$ matrix \quad \quad \quad \quad (b) Its frontal elimination tree or task-tree $T_0$ }   \label{M_11}
\label{Fig:Race}
\end{figure}

We first take a symmetric $11\times11$ matrix as an example.
Figure \ref{M_11}.(a) shows its pattern (only shows the lower triangle since it's symmetric). In this figure, non-zero element is represented by black dots  $\bullet$  and $\otimes$ is the extra fill-in. In the elimination process, there are some continuous rows(columns) with nearly same non-zero patterns, which is called frontal \cite{duff2017direct, irons1970frontal} or supernodal. These frontals are actually dense matrices, which can apply high performance BLAS operators. For the detail of frontal based method, please see \cite{duff2017direct, irons1970frontal, schenk2004solving,li2003superlu_dist,amestoy2000mumps,Chen2018}. 

There are many greedy algorithms to find the frontal structure. It's not the focus of this paper. We mainly study how to schedule these frontals to improve the efficiency of parallelism. For a frontal partition $[ \left \{ 1 \right \},\left \{ 2,3 \right \},\left \{ 4 \right \},\left \{ 5 \right \},\left \{ 6 \right \},\left \{ 7,9 \right \},\left \{ 8 \right \},\left \{ 11 \right \}]$, figure \ref{M_11}.(b) shows its frontal elimination tree\cite{gilbert1993elimination,liu1990role}. Each node of tree is a frontal, and the edges represent dependencies between frontals. We call this tree as the initial task-tree $T_0$, then study how to get different parallel task tree from $T_0$.

If two nodes have no dependence (neither parent node nor child node of each other), their elimination can be carried out at the same time. We could further extend it to all independent nodes. if there is no dependency between these nodes, their elimination can be carried out at the same time. We could divide these independent frontals into independent tasks.Then run these tasks on different cores to get high parallelism. As shown in Figure 3, there are three different task trees $T_1,T_2,T_3$ with different task partition. 
\begin{itemize}
    \item $T_1$: Delete edge $a,b$, get two parallel tasks=$[\{ 2,3 \},\{ 5 \},\{ 8 \}];[\{ 1 \},\{ 4 \},\{ 6 \},\{ 7,9 \}]$
    \item $T_2$: Delete edge $c,d,b$, get three parallel tasks=$[\{ 2,3 \}];[\{ 5 \}];[\{ 1 \},\{ 4 \},\{ 6 \},\{ 7,9 \}]$
    \item $T_3$: Delete edge $c,d,e,f$, get four parallel tasks=$[\{ 2,3 \}]; [\{ 5 \}]; [\{ 1 \}]; [\{ 4 \}]$
\end{itemize}
The three trees $T_1,T_2,T_3$ come from different delete actions on the initial task-tree $T_0$. We could also add some edges to get a new task-tree. For example, at the state of $T_2$, we could add edge  $c,d$ and re-delete edge $a$ to get $T_1$. We can try delete/add more edges to get more partitions!  How to get the optimal tree? We propose a novel Q-Learning based algorithm for this problem. Compared to the standard Q-Learning algorithm \ref{alg:Q} in section \ref{sec:Q}, different trees correspond to different states of MDP, and the action at each state is Delete/Add some edges. That is, the Q-Learning based task tree scheduling technique has three basic modules $\left ( T_n,\mathscr{A},R \right )$ :
\begin{enumerate}
\item State space:  $T_n$ is a set includes all possible split of task-tree.
\item Action space: $\mathscr{A}$=$\{ \textrm{Delete},\textrm{Add},\textrm{SKIP} \}$ 
    \begin{itemize}
        \item $\textrm{Delete}$: - Delete some edges to create more tasks. 
For example, at state $T_2$,  if some cores are idle, we could delete edge $e,f$ to get more tasks, then assign $[\{ 1 \}]; [\{ 4 \}]$ to idle cores. 
        \item $\textrm{Add}$: - Add some edges to union more tasks. 
For example, at state $T_3$,  we could delete edge $e,f$ to create a large task [\{ 1 \},\{ 4 \},\{ 6 \}]. 
        \item $\textrm{SKIP}$: The workloads of all computing cores are full. No need to change current task-tree.
    \end{itemize}
 
\item Reward function: $R$ represents the target of scheduling. The scheduling module needs to consider many factors. The main factors are listed below
\begin{itemize}
\item TIME - the wall time spent by the solver
\item MEMORY - memory consumption \cite{lacoste2015scheduling} 
\item BALANCE - workload balance balancing between the resources
\item OVERLAP - Efficient Communication/Computation Overlap \cite{marjanovic2010overlapping,sao2018communication}
\item POWER - Reduce the power and energy consumed\cite{aguilar2020performance}.
\end{itemize}
So the total reward is a  weighted sum of these metrics $R = \alpha \textrm{TIME} + \beta \textrm{MEM} + \gamma \textrm{BALANCE} + \cdots $.
\end{enumerate}

\begin{figure}[H]
\begin{subfigure}{\textwidth}
  \centering
  \includegraphics[width=0.7\textwidth]{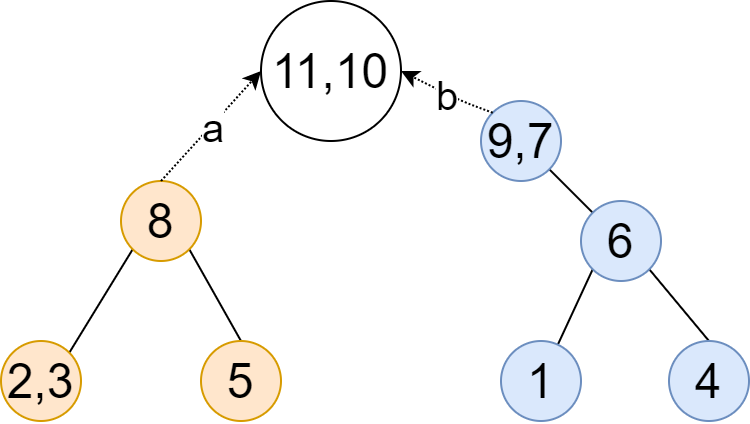}
  \caption{$T_1$: Delete edge $a,b$, get two parallel tasks=$[\{ 2,3 \},\{ 5 \},\{ 8 \}];[\{ 1 \},\{ 4 \},\{ 6 \},\{ 7,9 \}]$}
  \label{fig:sfig1}
\end{subfigure}

\begin{subfigure}{\textwidth}
  \centering
  \includegraphics[width=0.7\textwidth]{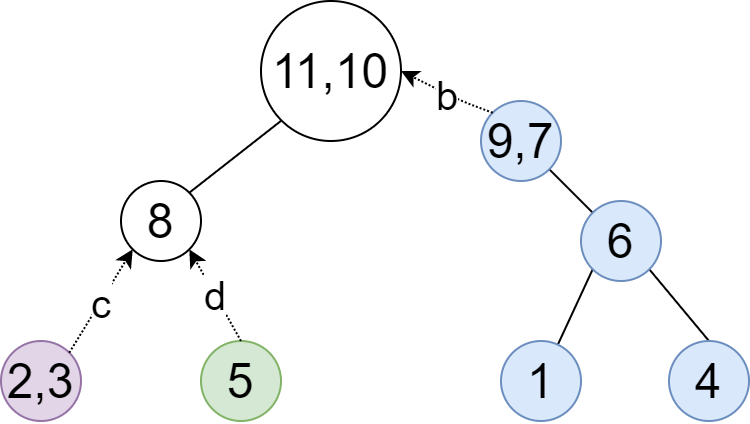}
  \caption{$T_2$: Delete edge $c,d,b$, get three parallel tasks=$[\{ 2,3 \}];[\{ 5 \}];[\{ 1 \},\{ 4 \},\{ 6 \},\{ 7,9 \}]$}
  \label{fig:sfig2}
\end{subfigure}

\begin{subfigure}{\textwidth}
  \centering
  \includegraphics[width=0.7\textwidth]{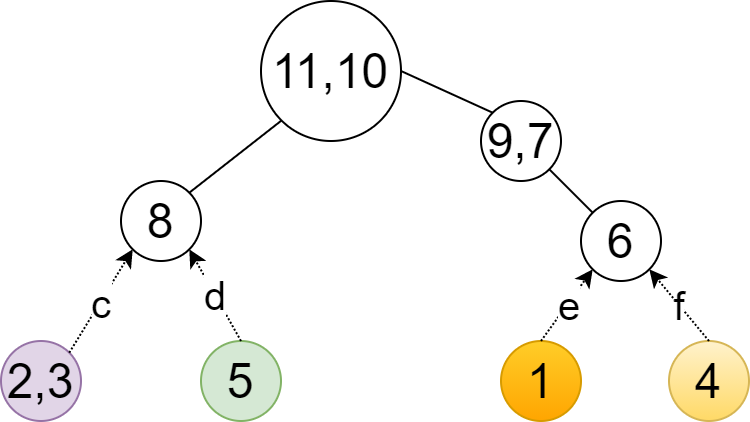}
  \caption{$T_3$: Delete edge $c,d,e,f$, get four parallel tasks=$[\{ 2,3 \}]; [\{ 5 \}]; [\{ 1 \}]; [\{ 4 \}]$}
  \label{fig:sfig3}
\end{subfigure}
\caption{ Different task partition on different split of task-tree} \label{fig:E-3}
\label{fig:fig}
\end{figure}

The following is the detailed algorithm to learn $Q$ (the expected rewards for an action taken in a given state $s$):
\begin{algorithm}[H]
\caption{An offline Q-Learning based task-tree scheduling technique} \label{alg:schedule-Q}
\begin{flushleft}
\textbf{Training stage:} 
    \\ \hspace*{0.2in} Learn $Q\left ( s,a \right )$ from the scheduling of many matrices(task-trees)
    \\ \hspace*{0.2in} 
    \begin{algorithmic} [1]
    \State Initialize $Q\left ( s,a \right )$. For the terminal state, $Q(terminal,\cdot)=0$
\For{ each episode}
    \State Pick a initial state $s$. It is just $T_0$(Elimination-Tree) of some matrix
    \While {true}
        \State Choose action $a$ for the current state $s$, using policy derived from $Q$ 
        \State Take action $a$ then enter the next state $s'$
        \State Get reward $r = R\left ( s,a \right )$
        \State Update Q by some method, for example:
        \State \quad $Q\left ( s,a \right ) \leftarrow Q\left ( s,a \right )+\alpha \left [ r+\gamma Q\left ( s',a' \right )-Q\left ( s,a \right )  \right ]$
        \State Set $s=s'$
        \If {$s$ is terminal}
            \State break
        \EndIf
    \EndWhile
\EndFor
    \end{algorithmic}

\textbf{Inference stage:}
\end{flushleft}
\begin{algorithmic} [1]
    \State The initial state $s$ is just $T_0$(Elimination-Tree)
    \While { un-eliminated frontals }
        \State For the current state $T_k$, take action $a$ from learned $Q$ table
        \State Run one or more task, eliminate some frontals, update task-tree
        \State Get reward $r = R\left ( T_k,p \right )$
        \State k=k+1
    \EndWhile
\end{algorithmic}
\end{algorithm}

\section{Experimental Result}
A new solver named ESS is developed using the above methods. We have implemented Q-Learning based scheduling algorithm in ESS. It uses Q-Learning based task-tree split technique to create an optimal task partition. Each task is assigned to a different computing core. In the actual elimination process, the task partition is based on pretrained $Q$ table. That is, if some cores have high computing power, then it will run more tasks automatically. 

To test the performance of ESS, 2 large equations with the ranks of 54603 and 126869 from power system unified dynamic simulations are taken for numeric factorization and solving. The CPU for testing is Intel i7-4770(3.4Ghz). The solver for comparison is KLU \cite{davis2010algorithm}, a direct serial sparse solver for circuit simulations, which is developed in Florida University. From the test results with 2~4 threads in parallel shown in table 2 and table 3, we can see that the speed-up of LU is about 1.9, 2.5 and 3.0 respectively and the speed-up of solving is about 1.5, 1.8 and 2.0. Also, the new algorithm in series is much faster than KLU.

\begin{figure}[ht]
\renewcommand{\figurename}{Table}
\begin{center}
\includegraphics[width=0.6\textwidth]{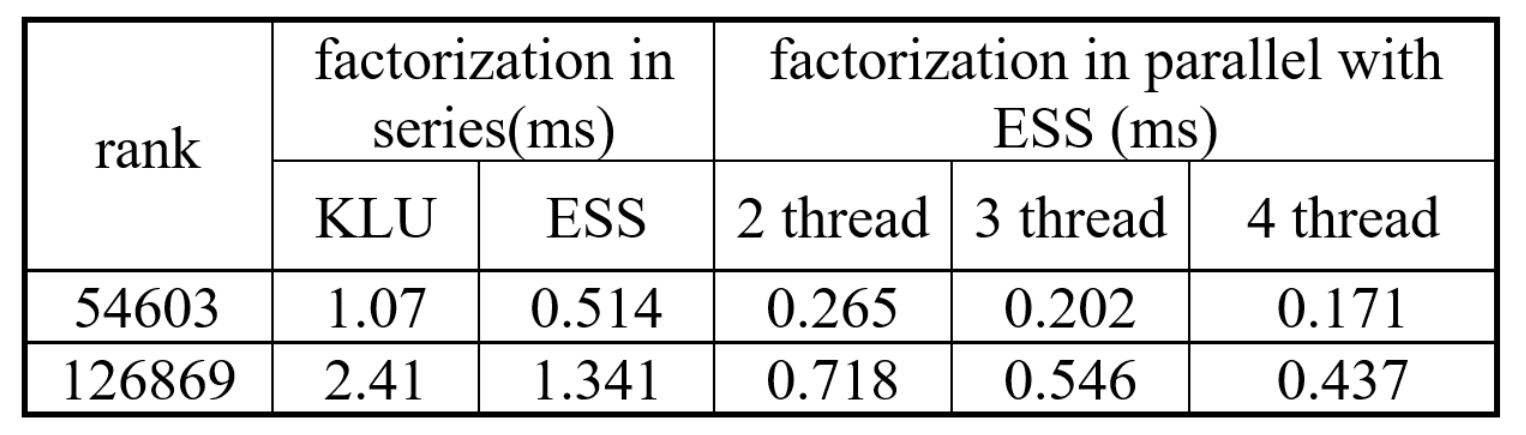}
\end{center}
\caption[position=bottom]{Time of numerical factorization(compared with KLU)}
\end{figure}

\begin{figure}[ht]
\renewcommand{\figurename}{Table}
\begin{center}
\includegraphics[width=0.6\textwidth]{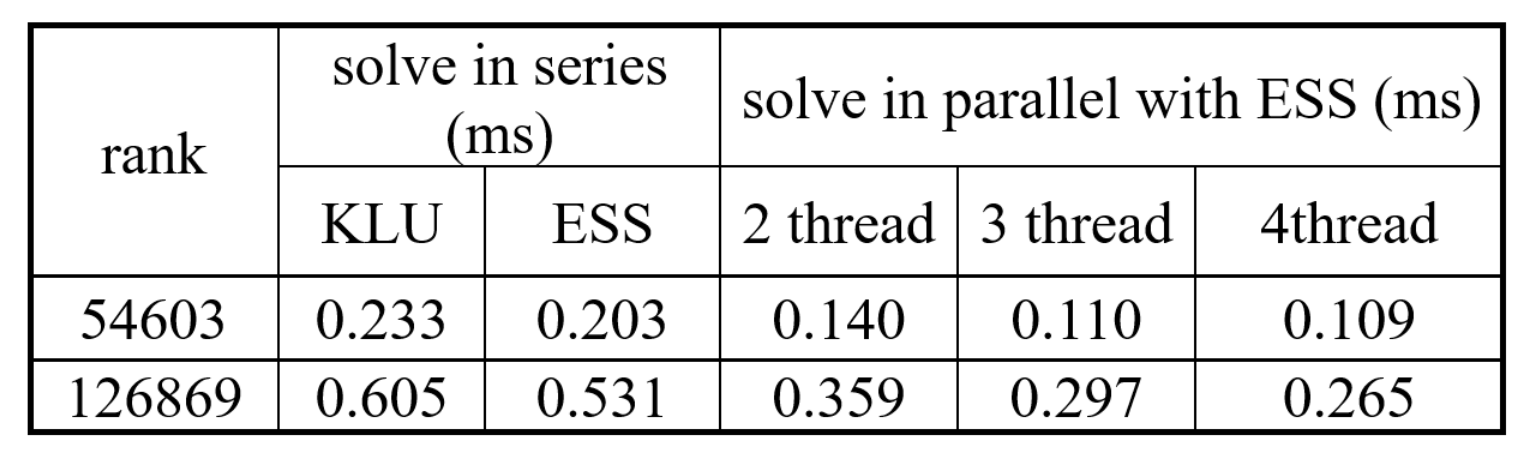}
\end{center}
\caption[position=bottom]{Time of solve(compared with KLU)}
\end{figure}
The test cases show that, with the Q-Learning adaptive scheduling technique, the new parallel algorithm can achieve higher computational efficiency on multi-core computers. 

\section{Conclusion}
We present a novel sparse solver with Q-Learning scheduler for the  dynamic simulations of power system. Numerical experiments on some large power systems shows that our solver is 2-6 times faster than classical sparse solver(KLU - the state-of-the-art sparse solver for circuit simulation problems). We would try more machine learning based method for this problem.

\bibliographystyle{unsrt}  
\bibliography{references}  






\end{document}